\pdfoutput=1

\documentclass[11pt]{article}

\usepackage[preprint]{acl}
\usepackage{xurl}
\usepackage{times}
\usepackage{latexsym}
\usepackage[T1]{fontenc}
\usepackage[utf8]{inputenc}
\usepackage{microtype}
\usepackage{inconsolata}
\usepackage{graphicx}

\title{Utilizing LLMs to Investigate the Disputed Role of Evidence in Electronic Cigarette Health Policy Formation in Australia and the UK}

\author{Damian Curran \\
  Computing \& Information Systems\\
  University of Melbourne \\
  Melbourne, Victoria, Australia \\
  \texttt{curran.d@unimelb.edu.au} \\\And
  Brian Chapman \\
  Computing \& Information Systems \\
  University of Melbourne \\
  Melbourne, Victoria, Australia\\
  \texttt{brian.chapman@unimelb.edu.au} \\
  \AND 
  Mike Conway\\
  Computing \& Information Systems\\
  University of Melbourne\\
  Melbourne, Victoria, Australia\\
  \texttt{mike.conway@unimelb.edu}}

\begin{document}
\maketitle
\begin{abstract}
Australia and the UK have developed contrasting approaches to the regulation of electronic cigarettes, with - broadly speaking - Australia adopting a relatively restrictive approach and the UK adopting a more permissive approach. Notably, these divergent policies were developed from the same broad evidence base.  In this paper, to investigate differences in how the two jurisdictions manage and present evidence, we developed and evaluated a Large Language Model-based sentence classifier to perform automated analyses of electronic cigarette-related policy documents drawn from official Australian and UK legislative processes (109 documents in total).  Specifically, we utilized GPT-4 to automatically classify sentences based on whether they contained claims that e-cigarettes were broadly helpful or harmful for public health.  Our LLM-based classifier achieved an F-score of 0.9.  Further, when applying the classifier to our entire sentence-level corpus, we found that Australian legislative documents show a much higher proportion of harmful statements, and a lower proportion of helpful statements compared to the expected values, with the opposite holding for the UK.  In conclusion, this work utilized an LLM-based approach to provide evidence to support the contention that - drawing on the same evidence base - Australian ENDS-related policy documents emphasize the harms associated with ENDS products and UK policy documents emphasize the benefits.  Further, our approach provides a starting point for using LLM-based methods to investigate the complex relationship between evidence and health policy formation.
\end{abstract}

\section{Introduction \& Background}

Australia and the United Kingdom (UK) have developed contrasting approaches to the regulation of Electronic Nicotine Delivery Systems (ENDS, also known as electronic cigarettes, e-cigs, and vapes).  UK policy development has been informed by a "harm reduction" approach in which existing adult smokers are encouraged to switch from combustible tobacco to (apparently, less harmful) ENDS use \citep{hawkinsStrategicUsesEvidence2019,smithContextualInfluencesRole2023}, while Australia has developed a more restrictive ENDS regulatory environment, motivated by concerns regarding the attractiveness of these products to young people and the risk of an unchecked market in ENDS creating a new generation of nicotine-dependent adults at risk of smoking-related harms \citep{freemanAustraliaReclaimingTobacco2023a,howeAustralianGovernmentsNew2024}. This divergence contrasts with the previously coordinated international approach to combustible tobacco regulation in the international tobacco control community.  For example the implementation of standardized health packaging \citep{hawkinsStrategicUsesEvidence2019}, the introduction of legislation stipulating a minimum age for purchasing tobacco products \citep{nuytsIncreaseTobaccoAgeofSale2020}, and the development of tax policies designed to limit tobacco consumption \citep{chaloupkaTobaccoTaxesTobacco2012}.

Notably, these diverse ENDS policies --- more permissive in the UK, more restrictive in Australia --- have developed despite public health bodies in the two jurisdictions drawing upon similar sources of evidence \citep{smithExaminingSourcesEvidence2021a,eisenkraftkleinUnderstandingExpertsConflicting2022}.  Several attempts have been made to cast light on this anomaly using a variety of approaches, including historical analyses of tobacco control approaches in the jurisdictions \citep{berridgeFirstPassUsing2021}, citation analysis of policy documents and position papers \citep{smithExaminingSourcesEvidence2021a}, and interviews with key informants in tobacco regulatory science policy \citep{smithEvidenceUseEcigarettes2021}.  Common themes that have emerged are the observation that Australian policy makers focus on the potential threats of ENDS products, while UK policy makers focus on potential benefits.  Several scholars have suggested that these policy differences are not necessarily entirely evidence-driven, but rather are influenced by underlying \emph{a priori} difference in policy orientations \citep{morphettDevelopmentEcigarettePolicy2023}.

In this work, we utilize Large Language Models (LLMs) to perform automated analyses of ENDS-related policy documents drawn from Australian and UK legislative processes. LLMs are complex neural network-based models that can be guided by prompt engineering to perform language processing tasks at scale. Both commercial and open-source LLMs are now commonly used in health applications. Examples include extracting adverse drug events from social media data \citep{liImprovingEntityRecognition2025}, developing and evaluating synthetic clinical text for training models in situations where data --- or data diversity --- is limited \citep{lozoyaGeneratingMentalHealth2024}, and performing information extraction from clinical notes \citep{holgateExtractingEpilepsyPatient2024}.  However, work on utilizing LLMs to better understand health policy formation is currently nascent.  

Our objective was to quantitatively analyze the divergent ENDS policy positions adopted in Australia and the UK. To do this, we quantified  differences in the framing of ENDS-related risks in public inquiry policy documents using an LLM-based, sentence-level classifier to label sentences as either \textit{helpful}, \textit{harmful}, or \textit{neither} with respect to the public health effects of ENDS,  evaluated this classifier on a manually annotated corpus of health policy statements, and - once validated - applied the classifier to a large collection of Australian and UK public policy documents, allowing us to identify clear differences in the interpretation of evidence between the two jurisdictions. Contributions of this work include (1) a robust empirical demonstration that Australian ENDS-related policy documents emphasize the harms associated with ENDS products and UK policy documents emphasize the benefits; and (2) a demonstration that LLMs can be utilized for work in health policy analysis, without the need to generate extensive training data as required by supervised language processing algorithms.

\section{Methods}

\subsection{Corpora}

In order to explore differences in evidential assertions between Australian and UK policy documents, we prepared three corpora for analysis, consisting of 109 documents (see Supplementary Materials\footnote{\url{https://maconway.github.io/ENDS_regulation.html}} for the corpora and source information.) The first corpus - henceforth, the \textbf{Erku corpus} - was originally compiled by \citet{erkuFramingScientificUncertainty2020a}, and consists of policy and position statements in Australia, the UK, and New Zealand on the subject of ENDS regulation. Given that our analysis is focused on Australia and the UK, we have excluded New Zealand-related documents.  The Erku corpus consisted of 36 documents after omitting the New Zealand documents and two documents that we were unable to locate.

We prepared two further corpora with a view to corroborating any findings from the Erku corpus.  The second corpus - the \textbf{Inquiry Documents (ID) corpus} - used written submissions made to public enquiries on ENDS policy in each jurisdiction, namely the 2018 House of Commons inquiry in the UK \citep{houseofcommonsScienceTechnologyCommittee2018} and the 2020 Senate inquiry in Australia \citep{australiansenateSelectCommitteeTobacco2020}. The third corpus - the \textbf{Inquiry Transcripts (IT) corpus} - consists of the official transcripts of the oral hearings from these public inquiries. Five oral hearings took place in the 2018 UK inquiry and two hearings took place as part of the 2020 Australian inquiry. In line with \citeposs{erkuFramingScientificUncertainty2020a} corpus creation strategy, we included reports from (and testimony on behalf of) government agencies, health charities, professional health bodies, and research groups, and excluded individuals presenting in their personal capacity, industrial lobby groups, and commercial entities. The ID corpus consisted of 47 documents (of which four documents also occur in the Erku corpus.) The IT corpus consisted of 26 transcripts. 

We converted all documents to a canonical plain text form.  Footnotes and references were deleted.  For the IT corpus, we included only the transcribed utterances spoken by the witness, not the questioner.  We used \texttt{SpaCy}\footnote{\url{https://spacy.io/}}  to split each text into its constituent sentences. In order to identify those sentences most likely to refer to evidence, only sentences that contained both a term related to ENDS (e.g. \textit{e-cig}, \textit{e cig}, \textit{ENDS}, \textit{electronic cigarette}, \textit{vape}) and a term related to evidence (e.g. \textit{evidence}, \textit{studies}, \textit{study}, \textit{research}, \textit{analysis}) were retained, resulting in a final dataset of 2,152 evidence sentences.

\subsection{Annotation Scheme}

\begin{table}[h] 
    \centering
    \begin{tabular}{p{5cm} c} 
        \hline
        \textbf{Sentence} & \textbf{Label} \\ 
        \hline
        Observational data from the UK suggest that there has been an increase in the popularity of e-cigarettes accompanied by a reduction in smoking cigarettes. & helpful \\
        \hline
        The Department considers that the evidence base supports maintaining and, where appropriate, strengthening the current controls that apply to the marketing and use of e-cigarettes in Australia. & harmful \\
        \hline
        It is clear that more research is needed, and that concerns about the potential for e-cigarettes to act as a gateway for non-smokers into tobacco smoking cannot be dismissed at this stage. & neither \\
        \hline
    \end{tabular}
    \caption{ENDS evidence sentence examples labelled with ternary annotation scheme labels}
    \label{tab:labeled_sentences}
\end{table}
We developed an annotation scheme to represent evidence-related statements regarding the risks and benefits of ENDS.  The three categories were \emph{helpful}, \emph{harmful}, or neither helpful nor harmful (\emph{neither}).  For instance, a sentence would be annotated as helpful if it stated that current evidence supported the hypothesis that ENDS products were useful for smoking cessation in some populations (e.g. "Survey results suggest that e-cigarettes represent an effective smoking cessation device among some respondents").  Conversely, a sentence would be annotated as harmful if it stated that there is evidence to support the position that ENDS devices encourage younger people to initiate smoking (e.g. "There is growing evidence that e-cigarette use is a precursor to smoking in young people").   Finally, the neither category encompasses sentences that do not make clear evidentiary claims regarding the harms or benefits of ENDS devices (e.g. "Research on e-cigarettes has been highlighted as a priority").  See \textbf{Table \ref{tab:labeled_sentences}} for more examples of manually labelled sentences.  Authors DC and MC manually labelled 200 randomly selected sentences according to the annotation scheme developed, with inter-annotator agreement (Cohen's kappa \citep{carlettaAssessingAgreementClassification1996a}) reported.  The annotation scheme developed for this work is reproduced in \textbf{Appendix \ref{appendix:a}}.

\subsection{Classification}

We used OpenAI's LLM, GPT-4 (gpt-4-0613), to automatically classify our annotated corpus of 200 sentences using a prompt template (i.e. static boilerplate text and an input slot which is updated on each iteration). In our case, the boilerplate text contained our annotation scheme and an instruction that the model should articulate its reasoning (as encouraging the model to explain its reasoning process has been demonstrated to improve classification performance for some reasoning tasks \citep{weiChainofthoughtPromptingElicits2022,liuPretrainPromptPredict2023}).  The input slot contained one of the 200 sentences to be classified (\textbf{Appendix \ref{appendix:b}} reproduces the prompt template used in this work).  After evaluating the performance of our LLM-based classifier on our annotated corpus of 200 sentences, we then applied the classifier to all 2,152 evidence sentences.  Finally, we compared the relative frequency of helpful and harmful ENDS-related evidence sentences as identified by our LLM-based classifier across the two jurisdictions using the chi-square statistic.

\section{Results}

\begin{table*}[h]
    \centering
    \begin{tabular}{llcccc} 
        \hline
        \textsc{Corpus Type} & \textsc{Country} & \# \textsc{Ev. Sent.} & \# \textsc{Helpful} & \# \textsc{Harmful} & \# \textsc{Neither} \\ \hline
        \textbf{Erku Corpus} & Australia & 218 & 7 (3\%) & 94 (43\%) & 117 \\
                                              & UK        & 848 & 159 (19\%) & 86 (10\%) & 603 \\
        \hline
        \textbf{ID Corpus}   & Australia & 560 & 78 (14\%) & 203 (36\%) & 279 \\
                                              & UK        & 384 & 118 (31\%) & 26 (7\%) & 240 \\
        \hline
        \textbf{IT Corpus}   & Australia & 88  & 17 (19\%) & 35 (40\%) & 36 \\
                                              & UK        & 54  & 11 (20\%) & 2 (4\%)   & 41 \\
        \hline
        \textbf{Total observed} & Australia & 866  & 102 (12\%) & 332 (38\%) & 432 \\
                                                 & UK        & 1,286 & 288 (22\%) & 114 (9\%)  & 884 \\
        \textbf{Expected values} & Australia & - & 202 & 231 & - \\
                                                  & UK        & - & 188 & 214 & - \\
        \hline
    \end{tabular}
    \caption{Summary of evidence sentence count and labels across corpus and country}
    \label{tab:evidence_summary}
\end{table*}
Inter-annotator agreement between the two annotators was high (kappa = 0.84), indicating that human annotators can reliably distinguish between helpful or harmful ENDS-related evidence claims.  Of the 200 sentences annotated, 36 were annotated as helpful, 35 as harmful, and 129 as neither.  Our prompt-based LLM-classifier achieved an F-score of 0.90 (precision, recall and accuracy all 0.90 when applied to the 2,152 evidence sentences).  The LLM-based classifier was then applied to our corpus of 2,152 sentences. We noted a clear difference between the proportions of \textit{helpful} and \textit{harmful} evidence statements compared to what would be expected if the distribution were independent of the country (measured using the chi-squared test, p < 0.0001). Australia shows a much higher proportion of harmful statements, and a lower proportion of helpful statements, compared to the expected values. The opposite was observed in the UK data. The labelling results are shown in \textbf{Table \ref{tab:evidence_summary}}.  Overall, our results clearly indicate that discussion regarding ENDS-related evidence in the UK policy context emphasises the potential benefits of ENDS for public health, whereas Australian policy documents place more emphasis on safety risks associated with ENDS devices. 

\section{Discussion}

We have empirically demonstrated that Australian ENDS-related policy documents and written and oral submissions to specific government inquiries emphasize the harms associated with ENDS products, whilst equivalent documents in the UK emphasize their benefits. These observations are consistent with Australia's restrictive policy approach to ENDS regulation, and the UK's more permissive policies.

Our work also serves as a demonstration of the utility of LLMs in health policy analysis. We were able to conduct highly accurate text analysis at scale, without the need to generate extensive labelled training data as required by traditional, supervised language processing algorithms. Further, our experimental framework could be modified to extract insights on, for example, target populations, commercial interests, tone, or appeals to underlying values within policy documents more generally.  To support reproducibility, code and data are provided in the Supplementary Materials.

\section{Limitations}
The work presented in this paper has several limitations.  First, our method for identifying ENDS-related evidence sentences may not have captured the full spectrum of ENDs-related evidence claims present in our dataset.  Second,  while we have shown that there are clear differences in the way that ENDS-related evidence is presented across the two jurisdictions, our results do not explain the underlying reason for these differences. Third, while we have taken steps to support scientific reproducibility (i.e. conducting an inter-annotator agreement study, publishing our data, code, and prompting instructions) given the probabilistic nature of LLMs, we cannot guarantee that precisely the same results would be achieved if the study where to be replicated.  Fourth, we have not experimented with alternative or baseline classification methods, given that the goal of the work is primarily a proof-of-concept demonstrating the utility of utilzing LLMs as a means of exploring the relationship between evidence and health policy formation.

\section{Conclusion}

This work demonstrates the feasibility of using LLMs to classify ENDS-related evidence sentences as either helpful (i.e. beneficial for public health) or harmful (i.e. detrimental for public health), with our LLM-based classifier achieving an F-score of 0.9, thus demonstrating that LLMs have a potentially useful role in health policy analysis.  Further, this work provides evidence to support the contention that - drawing on the same broad evidence base - Australian ENDS-related policy documents emphasize the harms associated with ENDS products and UK policy documents emphasize the benefits, illuminating the process by which the selection and interpretation of evidence is guided by underlying differences in policy orientation.

\section*{Ethics Statement}
The research reported in this paper is focused on the analysis of public domain policy and legislative documents and hence cannot be considered human subject research. 

\section*{Acknowledgments}
The research reported in this paper received no specific grant from any funding agency in the public, commercial, or not-for-profit sectors. 

\bibliography{ENDS_REGS}

\appendix

\section{Appendix: Annotation Scheme\label{appendix:a}}

Classify the sentence into one of these three categories: \textit{helpful}, \textit{harmful}, \textit{neither}. The categories correlate to what the
sentence asserts or implies about the health and other impact of e-cigarettes.

The assertion must be based on or implied from evidence, which includes studies, research, reports, finding, analysis, literature
or data.

The evidence must relate to e-cigarettes. E-cigarettes include e-cig, electronic cigarette, e-pen, EC, ENDS, vape and vaporizer. E-
cigarettes do not include tobacco cigarettes (also known as CC) or heat-not-burn products (also known as heated tobacco).  

\textit{Helpful} assertions or implications include (but are not limited to):

\begin{itemize}
\item e-cigarettes are better compared to tobacco smoking or cigarettes;
\item e-cigarettes reduce tobacco smoking or cigarette use;
\item e-cigarettes help people quit smoking.
\end{itemize}

\emph{Harmful} assertions or implications include (but are not limited to):

\begin{itemize}
\item e-cigarettes cause disease, illness, side-effects, addiction, fire, nicotine exposure or environmental harm;
\item e-cigarettes do not help people quit or reduce smoking;
\item e-cigarettes have the same negative effects as smoking;
\item e-cigarettes cause people to start smoking (the gateway effect);
\item e-cigarettes are used by people who never smoked tobacco previously (never smokers); and
\item e-cigarettes are used by children and adolescents (young people) (unless those young people were existing smokers).
\end{itemize}

The category is 'neither' if the sentence:

\begin{itemize}
\item contains both a helpful implication and a harmful implication;
\item is about the reason for using e-cigarettes;
\item is about the perception of e-cigarettes; or
\item is about the popularity or usage rates of e-cigarettes (except by young people and never-smokers).
\end{itemize}

The category is also 'neither' if the sentence states only that:
\begin{itemize}
\item the evidence is unclear;
\item the evidence is incomplete or inconclusive;
\item more evidence is needed; or
\item more evidence is being sought or was requested.
\end{itemize}

\section{Appendix: Prompt Template\label{appendix:b}}

Below is a sentence.

Give reasons and then classify the sentence into one of these three categories: ``helpful'', ``harmful'', ``neither''. The categories correlate to what the sentence asserts or implies about the health and other impacts of e-cigarettes.

The assertion must be based on or implied from evidence, which includes studies, research, reports, finding, analysis, literature or data.

The evidence must relate to e-cigarettes. E-cigarettes include e-cig, electronic cigarette, e-pen, EC, ENDS, vape and vaporizer. E-cigarettes do not include tobacco cigarettes (also known as CC) or heat-not-burn products (also known as heated tobacco).

``Helpful'' assertions or implications include (but are not limited to): e-cigarettes are better compared to tobacco smoking or cigarettes; e-cigarettes reduce tobacco smoking or cigarette use; and e-cigarettes help people quit smoking.

``Harmful'' assertions or implications include (but are not limited to): e-cigarettes cause disease, illness, side-effects, addiction, fire,
nicotine exposure or environmental harm; e-cigarettes do not help people quit or reduce smoking; e-cigarettes have the same
negative effects as smoking; e-cigarettes cause people to start smoking (the gateway effect); e-cigarettes are used by people who
never smoked tobacco previously (never smokers); e-cigarettes are used by children and adolescents (young people) (unless those
young people were existing smokers).

The category is ``neither'' if the sentence: contains both a helpful implication and a harmful implication; is about the reason for
using e-cigarettes; is about the perception of e-cigarettes; or is about the popularity or usage rates of e-cigarettes (except by
young people and never-smokers).

The category is also ``neither''' if the sentence states only that: the evidence is unclear; the evidence is incomplete or inconclusive;
more evidence is needed; or more evidence is being sought or was requested.

Format your reason and answer as follows: ``Reasoning:.... Answer:.....''

After ``Reasoning:'' provide concise reasoning for your answer. After ``Answer:'', put only one of these words: ``harmful'', ``helpful'' or
``neither''. Do nothing else.

This is the sentence: --> [INPUT SENTENCE]

\end{document}